\documentclass[11pt]{article}

\usepackage[final]{acl}

\usepackage{times}
\usepackage{latexsym}

\usepackage[T1]{fontenc}
\usepackage[utf8]{inputenc}

\usepackage{xcolor}

\usepackage{microtype}

\usepackage{inconsolata}
\usepackage{booktabs}
\usepackage{multirow}
\usepackage{graphicx}
\usepackage{lipsum,multicol}
\usepackage{tcolorbox}
\usepackage{enumitem}
\usepackage{subcaption}
\usepackage{hyperref}

\title{BLUCK: A Benchmark Dataset for Bengali Linguistic Understanding and Cultural Knowledge}

\author{
  Daeen Kabir\thanks{Equal contribution.}\thanks{Daeen Kabir conducted this research during his undergraduate studies at KAIST.}, 
  Minhajur Rahman Chowdhury Mahim\footnotemark[1]$^{\diamond}$, 
  Sheikh Shafayat$^{\diamond}$, \\
  \textbf{Adnan Sadik$^{\diamond}$, 
  Arian Ahmed$^{\diamond}$, 
  Eunsu Kim$^{\diamond}$, 
  Alice Oh$^{\diamond}$\thanks{Corresponding author}} \\[4pt]
  \footnotemark[2]Independent Researcher, $^{\diamond}$KAIST \\[4pt]
  \parbox{0.8\textwidth}{
    \centering
    \footnotesize
    \texttt{daeenkabirdk03@gmail.com} \\
    \texttt{\{minhaj, sheikh.shafayat, adnansadik235, arian.ahmed, kes0317\}@kaist.ac.kr} \\
    \texttt{alice.oh@kaist.edu}
  }
}

\begin{document}
\maketitle
\begin{abstract}

In this work, we introduce \textbf{BLUCK}, a new dataset designed to measure the performance of Large Language Models (LLMs) in Bengali linguistic understanding and cultural knowledge. Our dataset comprises 2366 multiple-choice questions (MCQs) carefully curated from compiled collections of several college and job level examinations and spans 23 categories covering knowledge on Bangladesh's culture and history and Bengali linguistics. We benchmarked BLUCK using 6 proprietary and 3 open-source LLMs - including GPT-4o, Claude-3.5-Sonnet, Gemini-1.5-Pro, Llama-3.3-70B-Instruct, and DeepSeekV3. Our results show that while these models perform reasonably well overall, they, however, struggles in some areas of Bengali phonetics. Although current LLMs' performance on Bengali cultural and linguistic contexts is still not comparable to that of mainstream languages like English, our results indicate Bengali's status as a mid-resource language. Importantly, BLUCK is also the first MCQ-based evaluation benchmark that is centered around native Bengali culture, history, and linguistics.
\end{abstract}

\section{Introduction}
% Recently, Large Language Models (LLMs), like the GPT and Claude series models, have been quite successful in solving various Natural Language Processing (NLP) tasks that range from question-answering tasks requiring general knowledge and commonsense to complex reasoning tasks. However, the showcase of these performances is limited to high-resource languages such as English or Chinese. Very recently, these LLMs have shown reasonable performance shown in multilingual settings for mid to low-resource languages; however, the evaluation has been limited to English-translated version of popular datasets like MMMLU (Massive Multitask Language Understanding) \cite{hendrycks2021measuringmassivemultitasklanguage}.

\begin{table}[ht]
\resizebox{\columnwidth}{!}{%
\begin{tabular}{@{}cc|c@{}}
\toprule
\multicolumn{2}{c|}{\textbf{Categories}}                                                   & \textbf{No. of Questions} \\ \midrule
\multicolumn{1}{c|}{\multirow{3}{*}{History}} & Ancient Bengal                  & 99                        \\
\multicolumn{1}{c|}{}                                    & British Bengal                  & 40                        \\
\multicolumn{1}{c|}{}                                    & Pakistan Era & 106                           
                \\ \midrule
\multicolumn{1}{c|}{\multirow{7}{*}{Culture}} & Indigenous People               & 31                        \\
\multicolumn{1}{c|}{}                                    & Arts, Heritage \& Media                             & 69                        \\
\multicolumn{1}{c|}{}                                    & National Issues                 & 15                        \\
\multicolumn{1}{c|}{}                                    & Constitution                    & 31                        \\
\multicolumn{1}{c|}{}                                    & Resources                       & 36                        \\
\multicolumn{1}{c|}{}                                    & Geography                       & 87                        \\
\multicolumn{1}{c|}{}                                    & Law                             & 284                       \\ \midrule
\multicolumn{1}{c|}{\multirow{7}{*}{Phonetics}}  & Alphabet                        & 10                        \\
\multicolumn{1}{c|}{}                                    & Pronunciation                   & 69                        \\
\multicolumn{1}{c|}{}                                    & Conjunct Letters                & 23                        \\
\multicolumn{1}{c|}{}                                    & Sound \& Letters               & 48                        \\
\multicolumn{1}{c|}{}                                    & Sound Changes                   & 54                        \\
\multicolumn{1}{c|}{}                                    & Phonetic Combining Rules        & 184                       \\
\multicolumn{1}{c|}{}                                    & Miscellaneous Phonetics         & 80                        \\ \midrule
\multicolumn{1}{c|}{\multirow{6}{*}{Semantics}}  & Synonyms                        & 364                       \\
\multicolumn{1}{c|}{}                                    & Antonyms                        & 165                       \\
\multicolumn{1}{c|}{}                                    & One Word Expressions            & 180                       \\
\multicolumn{1}{c|}{}                                    & Idioms                          & 198                       \\
\multicolumn{1}{c|}{}                                    & Proverbs                        & 47                        \\
\multicolumn{1}{c|}{}                                    & Miscellaneous                   & 146                       \\ \midrule
\multicolumn{2}{c|}{Total}                                                                 & 2366                      \\ \bottomrule
\end{tabular}%
}
\caption{Statistics of BLUCK}
\label{tab:my-table}
\end{table}
Recently, Large Language Models (LLMs) have demonstrated remarkable success in multilingual capabilities. In the case of Bengali, OpenAI’s O1 model achieved an impressive score of 0.873 \cite{openai2024openaio1card} on the MMLU benchmark. However, most evaluations of Bengali, including MMLU, rely on translated English datasets assessing general knowledge skills or focus exclusively on STEM fields, such as math and science~\cite{shafayat-etal-2024-benqa}. Despite the growing emphasis on evaluations that capture cultural and linguistic contexts for LLMs, the performance of models in Bengali-specific cultural knowledge or reasoning skills remains unexplored.

Given that Bengali is the 7th most spoken language in the world, with over 237 million native speakers, it is crucial to address the lack of high-quality Bengali-specific evaluation datasets. To this end, we introduce \textbf{BLUCK}\footnote{Dataset link: \href{https://github.com/minhaj1403/bluck/}{BLUCK}},
a Benchmark Dataset for Bengali Linguistic Understanding and
Cultural Knowledge. Through a rigorous curation process—encompassing careful annotation, multiple rounds of cross-inspection, and digitization—we have compiled a dataset of 2,366 multiple-choice questions (MCQs) that encompass extensive knowledge of the culture, history, and language of Bangladesh, into 23 subcategories. Table~\ref{tab:my-table} presents the overall statistics and categories of BLUCK.

Our evaluation of 9 LLMs using BLUCK offers valuable insights into the current status of LLMs in understanding Bengali language and cultural knowledge. Specifically, GPT-4o and Claude-3.5-Sonnet achieve the highest scores, around 73\% in a 0-shot setting—approximately 7\% lower than their performance on the MMLU benchmark. Overall, the models tend to perform well in the history category but show weaker results in the culture category, particularly on national issues. Similarly, in the phonetics category, their performance is generally low, with GPT-4o scores of 0.377 in pronunciation and 0.407 in sound changes. The lower performance in specific categories, such as culture and phonetics, highlights the current models' limitations in Bengali-specific knowledge. These findings underscore the potential for improvement in these areas, providing valuable insights for the future development of Bengali language models.

% Bengali is considered the 7th most spoken language in the world, with over 237 million native speakers. Despite its significance, there is a scarcity in the number of high-quality pretraining or evaluation datasets for Bengali \cite{mahfuz2024latetrainearlyuse}. Existing Bengali datasets are often created from English translations and fail to capture the native aspects of Bangladesh, and the Bengali language. To mitigate this biasness in representation of Bengali, we designed BLUCK to serve as a culturally and linguistically informed evaluation tool for LLMs. Following a rigorous curation process that involved careful annotation, multiple rounds of cross-inspection, and subsequent digitization, we compiled 2366 MCQs that include extensive knowledge on culture, history, and language of Bangladesh.\\

\begin{figure*}[t]
  \centering
  
  % --- Figure 1 ---
  % Set width to a fraction (e.g., 0.32\linewidth)
  \begin{subfigure}{\linewidth}
    \includegraphics[width=\linewidth]{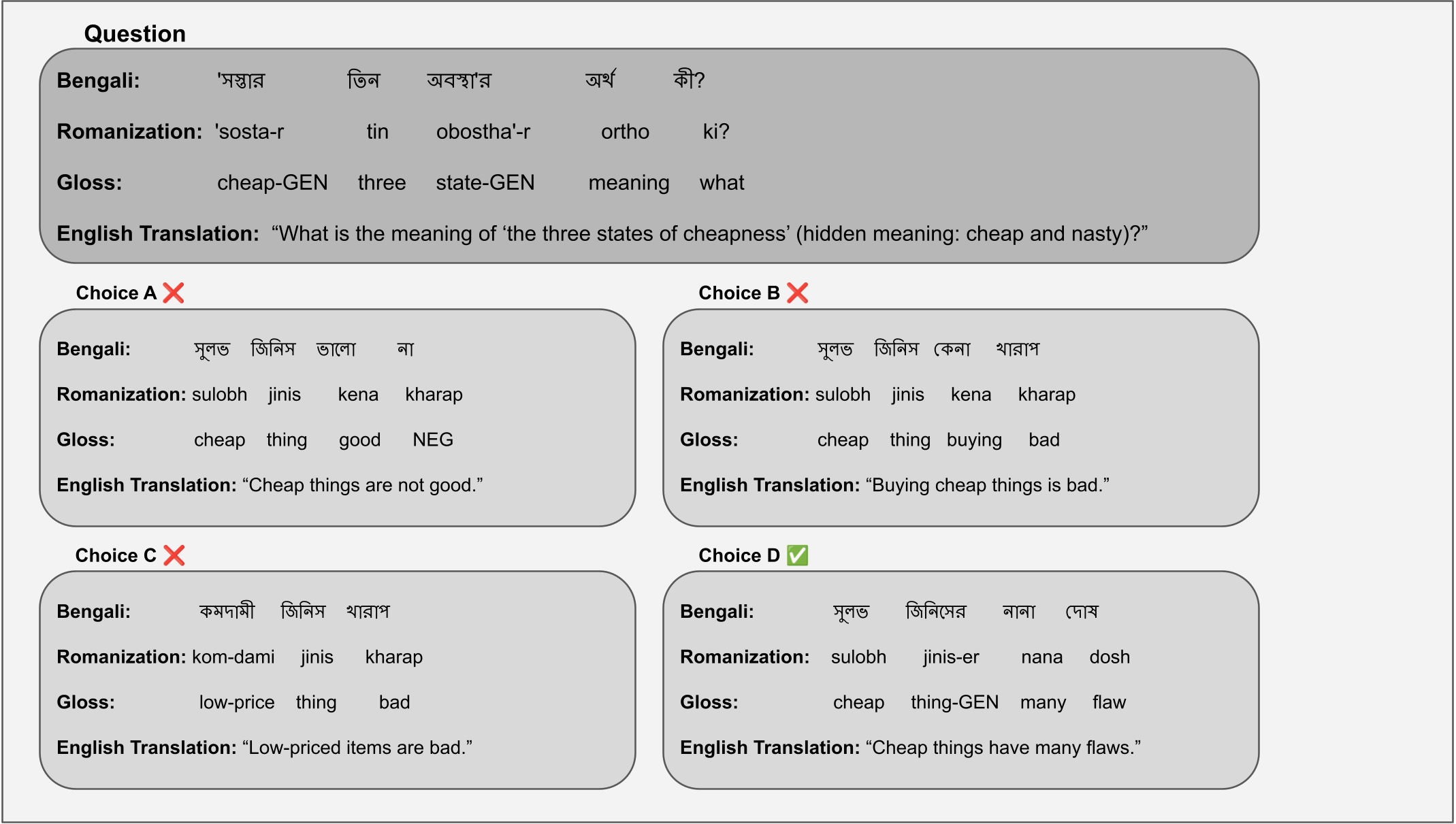}
    % \caption{Example MCQA from the Semantics domain.}
  \end{subfigure}
  % \hfill % Add horizontal space between 1 and 2
  \caption{Illustration of a MCQA sample from the Semantics domain. To aid linguistic understanding of non-native Bengali speakers, we include the original Bengali script, romanization, and word-by-word glossing, following the Leipzig Glossing Rules, alongside the English translation.}
\end{figure*}

\section{Related Works}

\subsection{Cultural Sensitive Dataset}
There has been a growing effort to create culturally sensitive benchmarks for evaluating LLMs across different languages and regions. Datasets like CULTURALBENCH \cite{chiu2024culturalbenchrobustdiversechallenging}, CLIcK \cite{kim-etal-2024-click} for Korean cultural knowledge RoCulturaBench \cite{masala2024vorbecstiromanecsterecipetrain} for Romanian culture are designed to assess LLMs' ability to understand cultural context beyond linguistic fluency. Similarly, BLEnD \cite{NEURIPS2024_8eb88844} provides a multilingual cultural dataset from more than 13 different languages. 

Despite these advances, Bengali remains significantly underrepresented in cultural-sensitive datasets. Most cultural evaluation benchmarks focus on high-resource languages or specific regional cultures, leaving a major gap in Bengali cultural and linguistic understanding.

\subsection{Bengali Dataset}
Several benchmarks have been developed to evaluate LLMs in general and multilingual tasks, but only a few focus on specific Bengali knowledge. M-MMLU \cite{openai2024mmmlu} and Global-MMLU \cite{singh2024globalmmluunderstandingaddressing} are some of the few well-known benchmarks that include Bengali in their multilingual evaluation settings. Nevertheless, their Bengali questions are mostly translated from English, limiting their effectiveness in assessing native-level linguistic understanding.

For Bengali reasoning tasks, MGSM \cite{shi2022language} provided the Bengali translations of GSM8K \cite{cobbe2021trainingverifierssolvemath}, one of the prominent datasets for grade school math problems along with other languages, although its scope remains limited. Later, BEnQA \cite{shafayat-etal-2024-benqa} provided multiple choice questions as official English-Bengali corpus, sourced from Bangladesh’s national board exams, focusing on STEM subjects. More recently, Tiger LLM \cite{raihan-zampieri-2025-tigerllm} aggregated diverse Bengali corpora, including educational, literary, and QA datasets, contributing significantly to Bengali NLP resources. Notably, there is still no phonetics-focused dataset for Bengali, which leaves granular but crucial aspects of the language absent from current evaluations. 

To the best of our knowledge, BLUCK is the first comprehensive benchmark to include Bengali-related reasoning and knowledge questions, focusing on Bengali history, culture, and language. BLUCK is specifically designed to evaluate LLMs in native Bengali contexts, complementing existing multilingual and subject-specific benchmarks. 

% , on the other hand, introduces a comprehensive benchmark for Bengali historical, cultural, and linguistic understanding. Unlike existing datasets, BLUCK is specifically designed to test LLMs in native Bengali contexts, complementing previous multilingual and subject-specific benchmarks.

\section{BLUCK: A  Benchmark Dataset for Bengali Linguistic Understanding and
Cultural Knowledge}
% \eunsu{recommend to write most of the part in ``present tense". FYI: chatgpt is good at converting these
% }
BLUCK contains immersive knowledge organized into these four major domains: Bangladesh's history, Bangladeshi culture, Bengali phonetics, and Bengali semantics. A summary of these categories, along with the corresponding number of questions is provided in Table~\ref{tab:my-table}.

\subsection{Data Collection}
Data is collected from publicly available printed copies of previous examination papers from the following sources: a) Bangladesh Civil Service (BCS) Examinations, b) university entrance examinations in Bangladesh, c) Bangladesh Bar Council Preliminary Examinations, d) bank job examinations, and e) several public job examinations. These official examinations are selected for their reliability and authoritative assessment of general knowledge in Bangladesh. These exams consist of extensive native knowledge on Bangladesh's history, culture, law, language, and various other academic disciplines. For BLUCK's creation, we select only the MCQs and follow a question selection criterion, based on which we omit these types of questions: a) fact-based questions loosely representing Bangladesh's history, culture, and language b) questions on contemporary issues in Bangladesh (to ensure long-term relevance), c) insignificant date-related or `number-based answer option' questions (to avoid arbitrary or trivial answers).

\subsection{Dataset Curation}
\paragraph{(1) Categorization}After data collection, we categorize by utilizing general knowledge and Bengali language guidebooks that organize questions similar to the ones in our dataset. This approach ensures proper categorization for some of the categories in the culture domain,  and allows us to group similar categories into the four main domains of our dataset.

\paragraph{(2) Two Round Inspection}The preliminary question selection task is distributed among the authors by category. To ensure quality, we implement two rounds of double-blind cross-checking. Independently, a second author collects questions for the same categories. The two authors then cross-check and agree on a preliminary set. In the second round, two more authors inspect these agreed-upon preliminary set in a double-blind manner, and cross-check their selection lists. All authors ensure that their selection process conforms to the aforementioned selection criteria . This process ensures that our dataset contains high-quality questions representing the history and culture of Bangladesh, and its rich linguistic knowledge.

\paragraph{(3) Digitization}After inspection, professional annotators, proficient in Bengali, digitize the MCQs for easier access and manipulation of the data. This is done to minimize the errors when digitized. To finalize our dataset, we conduct refinement: a) cleaning duplicate and inconsistent entries, b) correcting existing typing errors, and c) final checking to remove erroneous questions. This extensive approach ensures reliability and proper representation of the categories in our dataset.

\section{Experiment} 
\subsection{Experimental Setup}
In order to evaluate LLMs' performance on the history and culture of Bangladesh, and Bengali phonetics and semantics, we conduct experiments on the BLUCK dataset using both proprietary and open-source models. We utilized the following LLMs:

\begin{itemize}
    \item \textbf{Proprietary models}: GPT-4o, GPT-4o-mini~\cite{openai2024gpt4ocard}~\footnote{We use GPT-4o-2024-08-06 and GPT-4o-mini-2024-07-18 version using OPENAI API.}, Claude-3.5-Sonnet, Claude-3.5-Haiku~\footnote{We use Claude-3.5-Sonnet-20241022 and Claude-3.5-Haiku-20241022 version using Anthropic API.}, Gemini-1.5-Pro, Gemini-1.5-Flash~\cite{geminiteam2024gemini15unlockingmultimodal}~\footnote{We use gemini-1.5-pro-Latest, gemini-1.5-flash-latest using Gemini API key in Google AI Studio.}

    \item \textbf{Open-source models}: Llama-3.1-8B-Instruct, Llama-3.3-70B-Instruct~\cite{grattafiori2024llama3herdmodels}, DeepSeekV3~\cite{deepseekai2024deepseekv3technicalreport}
\end{itemize}

% Due to large-size of open-source models and limited computational resources, we use API-services for evaluation.
Since BLUCK consists largely of factual-knowledge based questions, we conduct evaluation without any chain-of-thought (CoT) reasoning, using both zero-shot and five-shot settings. As shown in Figure~\ref{fig:standard_prompt} in Appendix~\ref{sec:appendix}, for prompt, we utilize system and user prompts, explicitly instructing the model to output only the option `letter' in order to save API computational cost \cite{petrov2023token_unfairness}. Following the criteria in KoBBQ \cite{jin2024kobbqkoreanbiasbenchmark}, we only accept generated responses based on: a) response with only the alphabet as answer, b) response mentioning term corresponding to one of the options, iii) response convey the answer in the form `answer:', or `answer is', etc. Responses showing signs of hallucination, or producing bizarre outputs such as single Bengali letter as response are omitted.

% For the evaluation, we set maximum output token length to 1024 in order to save costs. and we set decoding temperature to 0.2 to reduce randomness, however, as shown in \cite{renze2024effectsamplingtemperatureproblem}, changing temperature from 0 to 1 do not have a significant performance change in LLMs. As for evaluation metric, we use accuracy.

\subsection{Result}
% \usepackage{multirow}
% \usepackage{graphicx}
\begin{table*}[t]
\centering
\resizebox{\textwidth}{!}{%
\begin{tabular}{cc|cc|cc|cc|cc|cc}
\hline
\multicolumn{2}{c|}{\multirow{2}{*}{Categories}}                               & \multicolumn{2}{c|}{GPT-4o}                          & \multicolumn{2}{c|}{Claude-3.5-Sonnet}       & \multicolumn{2}{c|}{Gemini-1.5-Pro}        & \multicolumn{2}{c|}{Llama-3.3-70B} & \multicolumn{2}{c}{DeepSeekV3}               \\
\multicolumn{2}{c|}{}                                                          & \multicolumn{1}{c|}{0-shot}         & 5-shot         & \multicolumn{1}{c|}{0-shot} & 5-shot         & \multicolumn{1}{c|}{0-shot} & 5-shot       & \multicolumn{1}{c|}{0-shot}  & 5-shot       & \multicolumn{1}{c|}{0-shot} & 5-shot         \\ \hline
\multicolumn{1}{c|}{\multirow{4}{*}{History}}       & Ancient Bengal           & \multicolumn{1}{c|}{0.899}          & \textbf{0.919} & \multicolumn{1}{c|}{0.879}  & 0.889          & \multicolumn{1}{c|}{0.758}  & 0.758        & \multicolumn{1}{c|}{0.687}   & 0.677        & \multicolumn{1}{c|}{0.859}  & 0.889          \\
\multicolumn{1}{c|}{}                               & British Bengal           & \multicolumn{1}{c|}{0.925}          & \textbf{0.975} & \multicolumn{1}{c|}{0.875}  & 0.9            & \multicolumn{1}{c|}{0.675}  & 0.95         & \multicolumn{1}{c|}{0.8}     & 0.85         & \multicolumn{1}{c|}{0.9}    & 0.95           \\
\multicolumn{1}{c|}{}                               & Pakistan Era             & \multicolumn{1}{c|}{0.745}          & \textbf{0.783} & \multicolumn{1}{c|}{0.67}   & 0.764          & \multicolumn{1}{c|}{0.481}  & 0.613        & \multicolumn{1}{c|}{0.5}     & 0.509        & \multicolumn{1}{c|}{0.717}  & 0.726          \\ \cmidrule{2-12}
\multicolumn{1}{c|}{}                               & \textbf{Average}            & \multicolumn{1}{c|}{0.837}          & \textbf{0.869} & \multicolumn{1}{c|}{0.788}   & 0.837          & \multicolumn{1}{c|}{0.624}  & 0.727        & \multicolumn{1}{c|}{0.624}     & 0.633        & \multicolumn{1}{c|}{0.804}  & 0.829          \\ \hline
\multicolumn{1}{c|}{\multirow{8}{*}{Culture}}       & Indigenous People        & \multicolumn{1}{c|}{0.806}          & 0.839          & \multicolumn{1}{c|}{0.871}  & \textbf{0.935} & \multicolumn{1}{c|}{0.516}  & 0.71         & \multicolumn{1}{c|}{0.516}   & 0.742        & \multicolumn{1}{c|}{0.774}  & 0.871          \\
\multicolumn{1}{c|}{}                               & Arts, Heritage \& Media  & \multicolumn{1}{c|}{0.739}          & \textbf{0.768} & \multicolumn{1}{c|}{0.725}  & 0.696          & \multicolumn{1}{c|}{0.58}   & 0.594        & \multicolumn{1}{c|}{0.58}    & 0.551        & \multicolumn{1}{c|}{0.725}  & \textbf{0.768} \\
\multicolumn{1}{c|}{}                               & National Issues          & \multicolumn{1}{c|}{0.467}          & 0.467          & \multicolumn{1}{c|}{0.733}  & \textbf{0.8}   & \multicolumn{1}{c|}{0.6}    & 0.733        & \multicolumn{1}{c|}{0.2}     & 0.6          & \multicolumn{1}{c|}{0.467}  & 0.6            \\
\multicolumn{1}{c|}{}                               & Constitution             & \multicolumn{1}{c|}{0.806}          & 0.871          & \multicolumn{1}{c|}{0.871}  & 0.935          & \multicolumn{1}{c|}{0.806}  & 0.903        & \multicolumn{1}{c|}{0.677}   & 0.71         & \multicolumn{1}{c|}{0.935}  & \textbf{0.968} \\
\multicolumn{1}{c|}{}                               & Resources                & \multicolumn{1}{c|}{0.778}          & 0.778          & \multicolumn{1}{c|}{0.722}  & 0.778          & \multicolumn{1}{c|}{0.5}    & 0.639        & \multicolumn{1}{c|}{0.528}   & 0.583        & \multicolumn{1}{c|}{0.694}  & \textbf{0.806} \\
\multicolumn{1}{c|}{}                               & Geography                & \multicolumn{1}{c|}{0.828}          & \textbf{0.862} & \multicolumn{1}{c|}{0.759}  & 0.793          & \multicolumn{1}{c|}{0.598}  & 0.724        & \multicolumn{1}{c|}{0.655}   & 0.621        & \multicolumn{1}{c|}{0.701}  & 0.77           \\
\multicolumn{1}{c|}{}                               & Law                      & \multicolumn{1}{c|}{0.68}           & 0.715          & \multicolumn{1}{c|}{0.648}  & \textbf{0.718} & \multicolumn{1}{c|}{0.613}  & 0.715        & \multicolumn{1}{c|}{0.496}   & 0.588        & \multicolumn{1}{c|}{0.588}  & 0.641          \\\cmidrule{2-12}
\multicolumn{1}{c|}{}                               & \textbf{Average}                  & \multicolumn{1}{c|}{0.725}          & \textbf{0.758} & \multicolumn{1}{c|}{0.707}  & \textbf{0.758}          & \multicolumn{1}{c|}{0.604}  & 0.707        & \multicolumn{1}{c|}{0.537}   & 0.604        & \multicolumn{1}{c|}{0.656}  & 0.718           \\ \hline
\multicolumn{1}{c|}{\multirow{8}{*}{Phonetics}}     & Alphabet                 & \multicolumn{1}{c|}{0.6}            & 0.7            & \multicolumn{1}{c|}{0.6}    & \textbf{0.9}   & \multicolumn{1}{c|}{0.2}    & \textbf{0.9} & \multicolumn{1}{c|}{0.6}     & \textbf{0.9} & \multicolumn{1}{c|}{0.6}    & \textbf{0.9}   \\
\multicolumn{1}{c|}{}                               & Pronunciation            & \multicolumn{1}{c|}{0.377}          & 0.406          & \multicolumn{1}{c|}{0.348}  & \textbf{0.507} & \multicolumn{1}{c|}{0.246}  & 0.391        & \multicolumn{1}{c|}{0.217}   & 0.333        & \multicolumn{1}{c|}{0.29}   & 0.348          \\
\multicolumn{1}{c|}{}                               & Conjunct Letters         & \multicolumn{1}{c|}{0.652}          & 0.739          & \multicolumn{1}{c|}{0.826}  & \textbf{0.957} & \multicolumn{1}{c|}{0.652}  & 0.826        & \multicolumn{1}{c|}{0.739}   & 0.826        & \multicolumn{1}{c|}{0.696}  & 0.826          \\
\multicolumn{1}{c|}{}                               & Sound \& Letters         & \multicolumn{1}{c|}{0.771}          & 0.729          & \multicolumn{1}{c|}{0.708}  & \textbf{0.792} & \multicolumn{1}{c|}{0.625}  & 0.771        & \multicolumn{1}{c|}{0.542}   & 0.688        & \multicolumn{1}{c|}{0.688}  & 0.75           \\
\multicolumn{1}{c|}{}                               & Sound Changes            & \multicolumn{1}{c|}{0.407}          & 0.611          & \multicolumn{1}{c|}{0.5}    & \textbf{0.667} & \multicolumn{1}{c|}{0.463}  & 0.63         & \multicolumn{1}{c|}{0.352}   & 0.537        & \multicolumn{1}{c|}{0.407}  & 0.574          \\
\multicolumn{1}{c|}{}                               & Phonetic Combining Rules & \multicolumn{1}{c|}{0.516}          & 0.603          & \multicolumn{1}{c|}{0.663}  & \textbf{0.761} & \multicolumn{1}{c|}{0.533}  & 0.609        & \multicolumn{1}{c|}{0.446}   & 0.473        & \multicolumn{1}{c|}{0.609}  & 0.63           \\
\multicolumn{1}{c|}{}                               & Miscellaneous Phonetics  & \multicolumn{1}{c|}{0.638}          & 0.675          & \multicolumn{1}{c|}{0.588}  & \textbf{0.7}   & \multicolumn{1}{c|}{0.5}    & 0.588        & \multicolumn{1}{c|}{0.463}   & 0.575        & \multicolumn{1}{c|}{0.575}  & 0.675          \\\cmidrule{2-12}
\multicolumn{1}{c|}{}                               & \textbf{Average}            & \multicolumn{1}{c|}{0.538}          & 0.609          & \multicolumn{1}{c|}{0.596}  & \textbf{0.718} & \multicolumn{1}{c|}{0.485}  & 0.609        & \multicolumn{1}{c|}{0.432}   & 0.526        & \multicolumn{1}{c|}{0.545}   & 0.618          \\ \hline
\multicolumn{1}{c|}{\multirow{7}{*}{Semantics}} & Synonyms                 & \multicolumn{1}{c|}{0.874}          & 0.912          & \multicolumn{1}{c|}{0.893}  & \textbf{0.923} & \multicolumn{1}{c|}{0.769}  & 0.835        & \multicolumn{1}{c|}{0.676}   & 0.772        & \multicolumn{1}{c|}{0.852}  & 0.907          \\
\multicolumn{1}{c|}{}                               & Antonyms                 & \multicolumn{1}{c|}{0.782}          & \textbf{0.891} & \multicolumn{1}{c|}{0.855}  & 0.879          & \multicolumn{1}{c|}{0.733}  & 0.812        & \multicolumn{1}{c|}{0.685}   & 0.739        & \multicolumn{1}{c|}{0.77}   & 0.848          \\
\multicolumn{1}{c|}{}                               & One Word Expressions     & \multicolumn{1}{c|}{0.717}          & 0.811          & \multicolumn{1}{c|}{0.778}  & 0.806          & \multicolumn{1}{c|}{0.589}  & 0.661        & \multicolumn{1}{c|}{0.556}   & 0.6          & \multicolumn{1}{c|}{0.717}  & \textbf{0.828} \\
\multicolumn{1}{c|}{}                               & Idioms                   & \multicolumn{1}{c|}{0.722}          & \textbf{0.808} & \multicolumn{1}{c|}{0.652}  & 0.747          & \multicolumn{1}{c|}{0.606}  & 0.662        & \multicolumn{1}{c|}{0.495}   & 0.505        & \multicolumn{1}{c|}{0.626}  & 0.697          \\
\multicolumn{1}{c|}{}                               & Proverbs                 & \multicolumn{1}{c|}{0.787}          & 0.83           & \multicolumn{1}{c|}{0.83}   & \textbf{0.894} & \multicolumn{1}{c|}{0.723}  & 0.809        & \multicolumn{1}{c|}{0.638}   & 0.745        & \multicolumn{1}{c|}{0.766}  & 0.787          \\
\multicolumn{1}{c|}{}                               & Miscellaneous            & \multicolumn{1}{c|}{\textbf{0.733}} & 0.712          & \multicolumn{1}{c|}{0.692}  & 0.719          & \multicolumn{1}{c|}{0.575}  & 0.589        & \multicolumn{1}{c|}{0.486}   & 0.514        & \multicolumn{1}{c|}{0.678}  & 0.719          \\ \cmidrule{2-12}
\multicolumn{1}{c|}{}                               & \textbf{Average}                 & \multicolumn{1}{c|}{0.785}          & \textbf{0.844} & \multicolumn{1}{c|}{0.795}  & 0.837          & \multicolumn{1}{c|}{0.677}  & 0.738        & \multicolumn{1}{c|}{0.598}   & 0.655        & \multicolumn{1}{c|}{0.750}   & 0.817          \\ \hline
\multicolumn{2}{c|}{Overall Average}                                                   & \multicolumn{1}{c|}{0.727}          & 0.780          & \multicolumn{1}{c|}{0.735}  & \textbf{0.795} & \multicolumn{1}{c|}{0.617}  & 0.704        & \multicolumn{1}{c|}{0.554}   & 0.615        & \multicolumn{1}{c|}{0.693}  & 0.756          \\ \hline
\end{tabular}%
}
\caption{BLUCK benchmark comparison by subcategories and major categories across major models in 0-shot and 5-shot settings. The highest accuracy(s) for each category are boldy marked.}
\label{tab:per-cat-results}
\end{table*}
Our evaluation results are summarized in Table {\textcolor{blue}{\ref{tab:per-cat-results}}}, which highlights the performance scores for all 23 categories of our dataset for the major models. Table {\textcolor{blue}{\ref{tab:per-cat-halfresults}}} in Appendix~\ref{sec:appendix} shows the same for the small-sized language models. Our results indicate that Claude-3.5-Sonnet, GPT-4o, Gemini-1.5-pro, and DeepSeekV3 demonstrate considerable knowledge of Bangladeshi history and the semantics of Bengali language. However, all the models struggle with phonetics, especially in areas such as pronunciation and sound changes. Claude-3.5-Sonnet emerges as the best overall model with consistent performance across all categories in both settings. It's performance in Bengali phonetics, which is the most difficult category, is 10\% better than the 2nd best model in this domain. GPT-4o closely follows, performing the best in history, culture, and semantics, while Claude-3.5-Sonnet achieves best performance in culture and phonetics.
The smaller models exhibit surprisingly reasonable performance, with Gemini-1.5-Flash and Claude-3.5-Haiku  surpassing even Llama-3.3-70B-Instruct in 5-shot setting. Llama-3.1-8B-Instruct, on the other hand, lags behind all other smaller models, showing very limited performance overall.

\section{Discussions}
% \eunsu{I think it's a short paper, so we don't need a thorough discussion.(and also, we don't have enough space). I think if the models got low scores in certain categories(e.g., linguistics if I remember correctly?) and discuss why this tendency(low accuracy) appears. you may also mention in which categories show low accuracy compared to mmlu score}

The benchmark results reveal significant variations in model performance across different categories and shot settings. Firstly, it is visible that 5-shot prompting leads to notable performance improvements (between 5\% to 10\%) across all models, which aligns with the findings that large language models pick up categorical cues from the examples and reduce the `search space' for MCQ solution under few-shot settings \cite{NEURIPS2020_1457c0d6}.

Secondly, proprietary models like GPT-4o and Claude-3.5-Sonnet consistently outperform open-source models for most of the categories, suggesting  that the former have a stronger contextual understanding of Bengali.

In addition, 'Pronunciation', and 'Sound Changes' are notable categories in which models exhibit poor performance. This strongly suggests that phonetic nuances in Bengali still remain underrepresented in existing LLMs, even with few-shot prompting.

The findings, overall, reinforce the need for more robust culture sensitive Bengali resources in LLM pretraining and evaluation benchmarks to improve performance in underrepresented Bengali linguistic and cultural areas.

\subsection{CoT Evaluation on Pronunciation}
We identify existing LLMs as performing markedly poor in the pronunciation category. To understand the source of this weakness, we conduct an additional experiment prompting Claude-3.5-Sonnet to use CoT reasoning on this particular subset under a zero-shot setting. Table~\ref{tab:pronunciation} shows the results.

The CoT approach did not yield considerable improvement. After evaluating the CoT responses, we attribute this mainly to insufficient knowledge of Bengali grammar as well as limited understanding of Bengali phonetic rules. In several cases, the model made an error in the very first reasoning step due to incorrect background assumptions, causing the entire reasoning chain to collapse. We also observed a tendency for the model to fixate on an answer prematurely and then generate a forced rationale to justify it. Cases are listed in the Appendix \ref{sec:appendix}.

\begin{table}[h]
\centering
\small
\begin{tabular}{l|ccc}
\hline
\textbf{Pronunciation} & \multicolumn{3}{c}{Claude 3.5-Sonnet} \\
\cline{2-4}
 & 0-shot & 5-shot & \textbf{0-shot CoT} \\
\hline
Accuracy & 0.348 & 0.507 & 0.478 \\
\hline
\end{tabular}
\caption{Claude 3.5 performance on the Pronunciation subcategory.}
\label{tab:pronunciation}
\end{table}

% We observe an increase of about between 5\% to 10\% for all models 
% from 0-shot setting to 5-shot setting. One particular reason is that large language models are few-shot learners \cite{NEURIPS2020_1457c0d6}, which enables models to pick up categorical cues from the examples and reduce the `search space' for MCQ solution.
\section{Conclusion}
We introduced BLUCK, a linguistic and culture-sensitive Bengali dataset, locally sourcing from official college and job-level examinations in Bangladesh. BLUCK provides a diverse set of 2366 multiple-choice questions that fall under 23 subcategories organized across four domains. Our evaluation using state-of-the-art LLMs showcases their knowledge in historical and semantics aspects of Bengali, while exposes their weakness in linguistically nuanced areas. Future research should expand BLUCK and improve LLMs' understanding of Bengali linguistic and cultural nuances. 

% In this work, we introduce a novel evaluation suite to measure LLMs' prowess in Bangladesh-specific cultural and historical knowledge, and Bengali linguistic knowledge. Sourced from official college and job-level examinations, BLUCK provides a diverse set of 2366 multiple-choice questions that fall under 23 subcategories organized across four domains. Our evaluation using state-of-the-art LLMs showcases their knowledge in historical and semantics aspects of Bengali, while exposes their weakness in culturally and linguistically nuanced areas, particularly in phonetics and national issues.

\section*{Limitations}
We acknowledge certain limitations in our work. Since our dataset consists solely of text-based questions, we cannot determine whether the models arrived at their answers through reasoning processes different from those of humans. Moreover, given the richness of Bengali culture, history, and linguistic diversity, as well as the growing importance of M-MMLU, Global-MMLU and other large-scale multilingual benchmarks, our contribution remains relatively small in comparison. However, we hope that BLUCK serves as a stepping stone to improve Bengali culture-sensitive LLM research.

\section*{Ethical Considerations}

The BLUCK dataset is fully available and has been manually curated and reviewed to mitigate any chance of having harmful contents. This dataset will be publicly accessible and distributed under the CC BY-SA 4.0 license. Our work has been reviewed and received approval from the Institutional Review Board (IRB) at our institution. All annotators involved in this project were compensated above the minimum wage and standards. Finally, AI-assisted tools were used solely for grammar and language refinement. They were not used for writing, analysis, or coding in any capacity.

% \section*{Acknowledgements}
% XYZ

% The exam questions in BEnQA dataset are freely 543
% available and have been manually curated and re- 544
% viewed to minimize the presence of harmful con- 545
% tent. This dataset will be publicly accessible and 546
% distributed under the CC BY-SA 4.0 license. Our 547
% work has been reviewed and received approval from 548
% the Institutional Review Board (IRB) at our institu- 549
% tion. All annotators involved in this project were 550
% compensated above the minimum wage.


%case for working on and arr review copy
\bibliography{custom}

%case for arxiv
% \input{acl_latex.bbl}

\clearpage
\section{Appendix}
\label{sec:appendix}
\subsection{Evaluation Details}

Since our MCQ questions are largely factual-based and do not require reasoning for most cases, we set the maximum output token length is set to 1024 for all experiments. This allows use to analyze responses from models during cases where models produce verbose responses, primarily in 0-shot setting, due to lack of guiding examples in 5-shot setting, despite being instructed in the prompt to produce only option ID as output. We set the decoding temperature to 0.2 to reduce randomness, however, as shown in \cite{renze2024effectsamplingtemperatureproblem}, changing temperature from 0 to 1 do not have a significant performance change in LLMs.

For 5-shot setting, we randomly pick 5 questions from each category. Since we perform a meticulous categorization and double-inspection process, our randomly selected samples are generally good representations of the category.

\begin{figure}[h]
  \includegraphics[width=\linewidth]{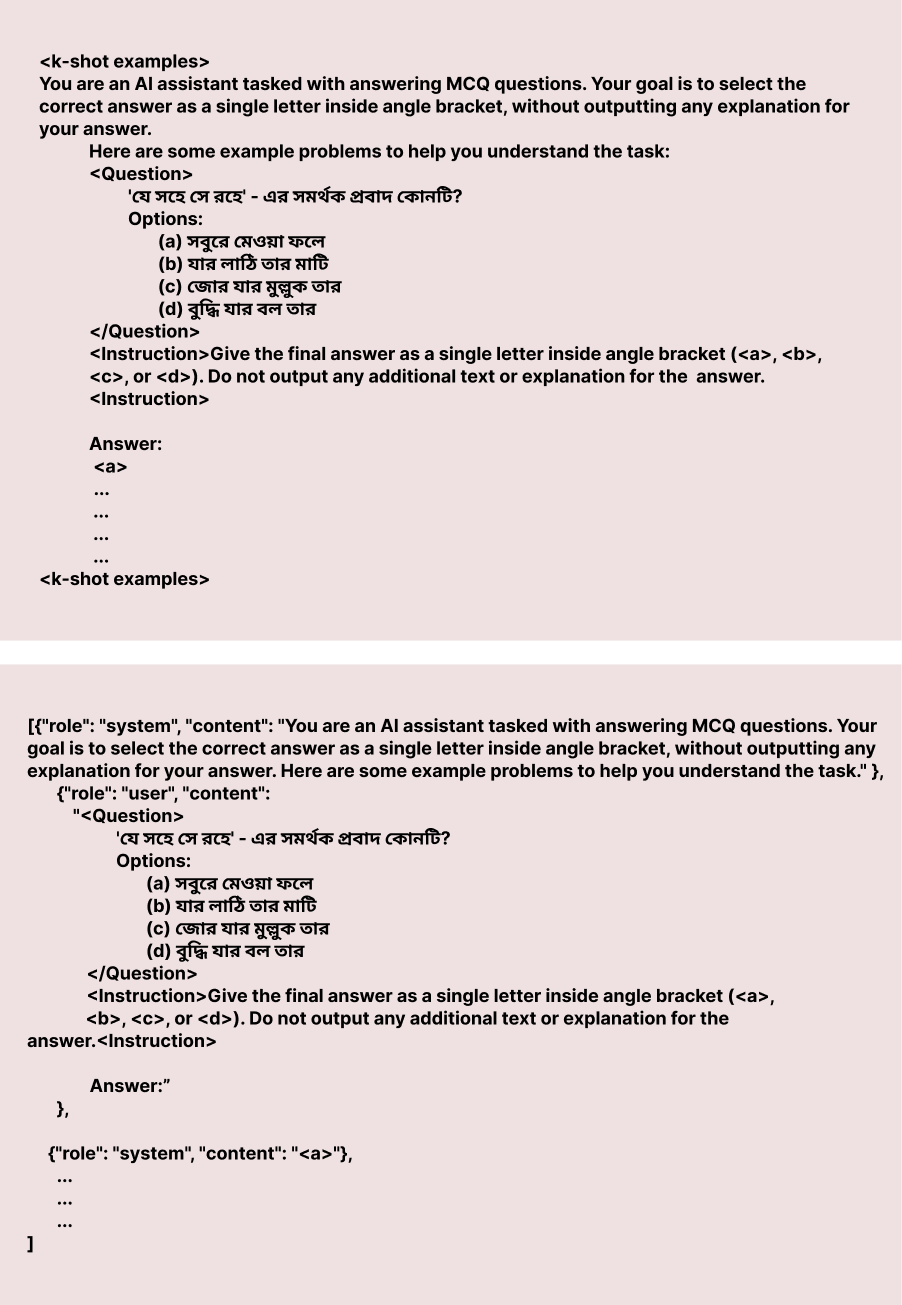} 
  \caption {Prompt Structure for 5-shot setting using GPT model.}
  \label{fig:gptclaudeprompt}
\end{figure}
% --- START OF COMBINED FIGURES ---
\begin{figure}[ht]
    \centering
    % First Subfigure (Left)
    \begin{subfigure}[b]{0.48\textwidth}
        \centering
        \includegraphics[width=\linewidth]{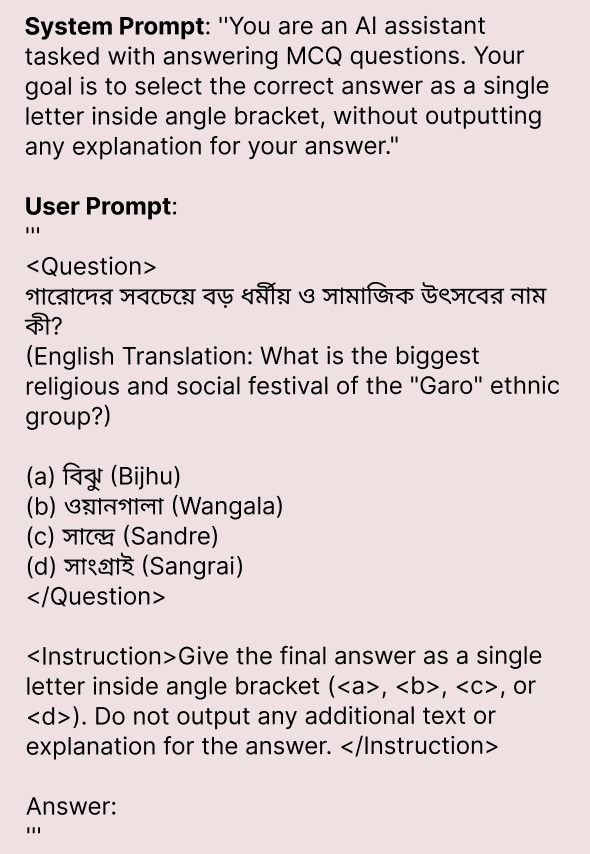}
        \caption{Illustration of our zero-shot standard prompt.}
        \label{fig:standard_prompt}
    \end{subfigure}
    \hfill % Adds space between the images
    % Second Subfigure (Right)
    \begin{subfigure}[b]{0.48\textwidth}
        \centering
        \includegraphics[width=\linewidth]{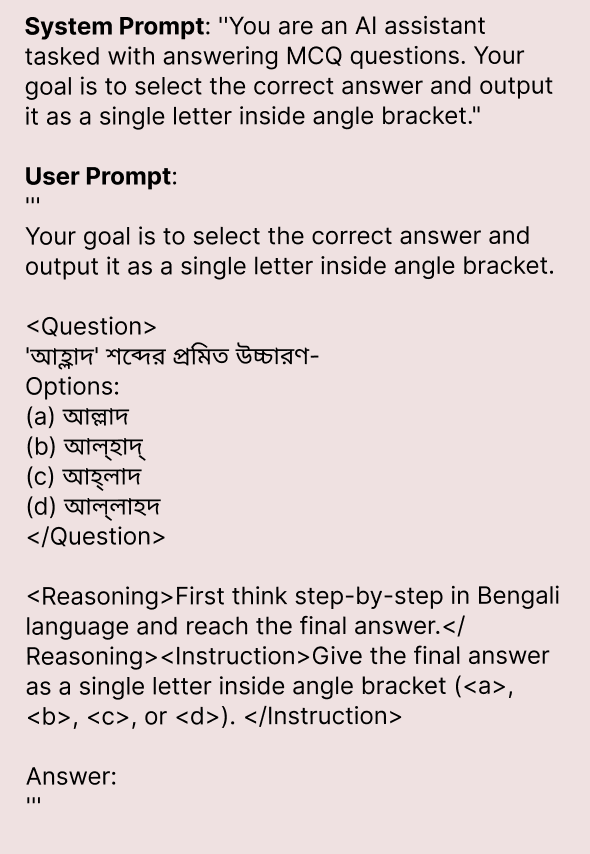}
        \caption{Prompt of zero-shot CoT reasoning by Claude-3.5-Sonnet on the Pronunciation category.}
        \label{fig:cot_prompt}
    \end{subfigure}
    
    \caption{Prompting strategies used in our evaluation.}
    \label{fig:prompt_strategies}
\end{figure}
% % --- END OF COMBINED FIGURES ---

% \begin{figure}[h]
%   \includegraphics[width=\columnwidth]{Diagrams/prompt.png} 
%   \caption {Illustration of our prompt.}
%   \label{fig:appendix_figure}
% \end{figure}

% \newpage
% \textbf{Prompting Strategies} 
% \begin{figure}[h]
%   \includegraphics[width=\linewidth]{Diagrams/CoT_Prompt.png} 
%   \caption {Prompt of zero-shot CoT reasoning by Claude-3.5-Sonnet on the Pronunciation category.}
%   \label{fig:gptclaudeprompt}
% \end{figure}

\subsection{Additional BLUCK Results}

% \usepackage{multirow}
% \usepackage{graphicx}
\begin{table*}[ht]
\centering
\resizebox{\textwidth}{!}{%
\begin{tabular}{cc|cc|cc|cc|cc}
\hline
\multicolumn{2}{c|}{\multirow{2}{*}{Categories}}                               & \multicolumn{2}{c|}{GPT-4o-mini}                     & \multicolumn{2}{c|}{Claude-3.5-Haiku}        & \multicolumn{2}{c|}{Gemini-1.5-Flash}        & \multicolumn{2}{c}{Llama-3.1-8B} \\
\multicolumn{2}{c|}{}                                                          & \multicolumn{1}{c|}{0-shot}         & 5-shot         & \multicolumn{1}{c|}{0-shot} & 5-shot         & \multicolumn{1}{c|}{0-shot} & 5-shot         & \multicolumn{1}{c|}{0-shot}    & 5-shot   \\ \hline
\multicolumn{1}{c|}{\multirow{4}{*}{History}}       & Ancient Bengal           & \multicolumn{1}{c|}{0.667}          & \textbf{0.737} & \multicolumn{1}{c|}{0.687}  & 0.717          & \multicolumn{1}{c|}{0.646}  & 0.697          & \multicolumn{1}{c|}{0.394}     & 0.404    \\
\multicolumn{1}{c|}{}                               & British Bengal           & \multicolumn{1}{c|}{0.775}          & 0.8            & \multicolumn{1}{c|}{0.7}    & \textbf{0.825} & \multicolumn{1}{c|}{0.675}  & 0.8            & \multicolumn{1}{c|}{0.35}      & 0.475    \\
\multicolumn{1}{c|}{}                               & Pakistan Era             & \multicolumn{1}{c|}{0.528}          & \textbf{0.566} & \multicolumn{1}{c|}{0.453}  & 0.491          & \multicolumn{1}{c|}{0.377}  & 0.491          & \multicolumn{1}{c|}{0.302}     & 0.396    \\
\multicolumn{1}{c|}{}                               & Average             & \multicolumn{1}{c|}{0.624}          & \textbf{0.673} & \multicolumn{1}{c|}{0.588}  & 0.637          & \multicolumn{1}{c|}{0.535}  & 0.624          & \multicolumn{1}{c|}{0.347}     & 0.412    \\ \hline
\multicolumn{1}{c|}{\multirow{8}{*}{Culture}}       & Indigenous People        & \multicolumn{1}{c|}{0.484}          & \textbf{0.677} & \multicolumn{1}{c|}{0.452}  & 0.645          & \multicolumn{1}{c|}{0.355}  & 0.581          & \multicolumn{1}{c|}{0.355}     & 0.452    \\
\multicolumn{1}{c|}{}                               & Arts, Heritage \& Media  & \multicolumn{1}{c|}{0.478}          & \textbf{0.536} & \multicolumn{1}{c|}{0.42}   & 0.478          & \multicolumn{1}{c|}{0.449}  & 0.449          & \multicolumn{1}{c|}{0.29}      & 0.377    \\
\multicolumn{1}{c|}{}                               & National Issues          & \multicolumn{1}{c|}{0.4}            & 0.533          & \multicolumn{1}{c|}{0.267}  & 0.4            & \multicolumn{1}{c|}{0.467}  & \textbf{0.667} & \multicolumn{1}{c|}{0.133}     & 0.533    \\
\multicolumn{1}{c|}{}                               & Constitution             & \multicolumn{1}{c|}{0.677}          & 0.774          & \multicolumn{1}{c|}{0.581}  & 0.71           & \multicolumn{1}{c|}{0.645}  & \textbf{0.839} & \multicolumn{1}{c|}{0.355}     & 0.387    \\
\multicolumn{1}{c|}{}                               & Resources                & \multicolumn{1}{c|}{0.472}          & \textbf{0.722} & \multicolumn{1}{c|}{0.417}  & 0.5            & \multicolumn{1}{c|}{0.472}  & 0.583          & \multicolumn{1}{c|}{0.25}      & 0.389    \\
\multicolumn{1}{c|}{}                               & Geography                & \multicolumn{1}{c|}{0.563}          & \textbf{0.598} & \multicolumn{1}{c|}{0.46}   & 0.506          & \multicolumn{1}{c|}{0.471}  & 0.529          & \multicolumn{1}{c|}{0.299}     & 0.333    \\
\multicolumn{1}{c|}{}                               & Law                      & \multicolumn{1}{c|}{0.472}          & 0.546          & \multicolumn{1}{c|}{0.496}  & 0.514          & \multicolumn{1}{c|}{0.493}  & \textbf{0.577} & \multicolumn{1}{c|}{0.345}     & 0.405    \\
\multicolumn{1}{c|}{}                               & Average             & \multicolumn{1}{c|}{0.497}          & \textbf{0.584} & \multicolumn{1}{c|}{0.472}  & 0.523          & \multicolumn{1}{c|}{0.483}  & 0.571          & \multicolumn{1}{c|}{0.320}     & 0.394    \\ \hline
\multicolumn{1}{c|}{\multirow{8}{*}{Phonetics}}     & Alphabet                 & \multicolumn{1}{c|}{0.2}            & 0.8            & \multicolumn{1}{c|}{0.7}    & 0.8            & \multicolumn{1}{c|}{0.7}    & \textbf{0.9}   & \multicolumn{1}{c|}{0.2}       & 0.7      \\
\multicolumn{1}{c|}{}                               & Pronunciation            & \multicolumn{1}{c|}{0.159}          & 0.275          & \multicolumn{1}{c|}{0.261}  & 0.275          & \multicolumn{1}{c|}{0.203}  & \textbf{0.319} & \multicolumn{1}{c|}{0.217}     & 0.29     \\
\multicolumn{1}{c|}{}                               & Conjunct Letters         & \multicolumn{1}{c|}{0.478}          & 0.652          & \multicolumn{1}{c|}{0.783}  & \textbf{0.957} & \multicolumn{1}{c|}{0.609}  & 0.783          & \multicolumn{1}{c|}{0.478}     & 0.522    \\
\multicolumn{1}{c|}{}                               & Sound \& Letters         & \multicolumn{1}{c|}{0.5}            & 0.625          & \multicolumn{1}{c|}{0.438}  & 0.646          & \multicolumn{1}{c|}{0.625}  & \textbf{0.667} & \multicolumn{1}{c|}{0.25}      & 0.292    \\
\multicolumn{1}{c|}{}                               & Sound Changes            & \multicolumn{1}{c|}{0.278}          & 0.333          & \multicolumn{1}{c|}{0.315}  & 0.444          & \multicolumn{1}{c|}{0.278}  & \textbf{0.481} & \multicolumn{1}{c|}{0.296}     & 0.352    \\
\multicolumn{1}{c|}{}                               & Phonetic Combining Rules & \multicolumn{1}{c|}{0.402}          & 0.418          & \multicolumn{1}{c|}{0.435}  & \textbf{0.505} & \multicolumn{1}{c|}{0.457}  & 0.478          & \multicolumn{1}{c|}{0.31}      & 0.359    \\
\multicolumn{1}{c|}{}                               & Miscellaneous Phonetics  & \multicolumn{1}{c|}{0.55}           & 0.6            & \multicolumn{1}{c|}{0.525}  & \textbf{0.613} & \multicolumn{1}{c|}{0.575}  & 0.575          & \multicolumn{1}{c|}{0.4}       & 0.363    \\
\multicolumn{1}{c|}{}                               & Average         & \multicolumn{1}{c|}{0.387}          & 0.459          & \multicolumn{1}{c|}{0.434}  & \textbf{0.526} & \multicolumn{1}{c|}{0.449}  & 0.515          & \multicolumn{1}{c|}{0.310}     & 0.357    \\ \hline
\multicolumn{1}{c|}{\multirow{7}{*}{Semantics}} & Synonyms                 & \multicolumn{1}{c|}{0.681}          & 0.747          & \multicolumn{1}{c|}{0.761}  & \textbf{0.843} & \multicolumn{1}{c|}{0.687}  & 0.775          & \multicolumn{1}{c|}{0.385}     & 0.426    \\
\multicolumn{1}{c|}{}                               & Antonyms                 & \multicolumn{1}{c|}{0.642}          & 0.691          & \multicolumn{1}{c|}{0.679}  & \textbf{0.77}  & \multicolumn{1}{c|}{0.691}  & 0.758          & \multicolumn{1}{c|}{0.412}     & 0.527    \\
\multicolumn{1}{c|}{}                               & One Word Expressions     & \multicolumn{1}{c|}{0.506}          & 0.567          & \multicolumn{1}{c|}{0.589}  & \textbf{0.661} & \multicolumn{1}{c|}{0.522}  & 0.606          & \multicolumn{1}{c|}{0.433}     & 0.406    \\
\multicolumn{1}{c|}{}                               & Idioms                   & \multicolumn{1}{c|}{0.515}          & 0.5            & \multicolumn{1}{c|}{0.444}  & 0.495          & \multicolumn{1}{c|}{0.48}   & \textbf{0.581} & \multicolumn{1}{c|}{0.354}     & 0.333    \\
\multicolumn{1}{c|}{}                               & Proverbs                 & \multicolumn{1}{c|}{0.66}           & 0.66           & \multicolumn{1}{c|}{0.638}  & 0.723          & \multicolumn{1}{c|}{0.66}   & \textbf{0.787} & \multicolumn{1}{c|}{0.404}     & 0.426    \\
\multicolumn{1}{c|}{}                               & Miscellaneous            & \multicolumn{1}{c|}{\textbf{0.616}} & 0.589          & \multicolumn{1}{c|}{0.521}  & 0.568          & \multicolumn{1}{c|}{0.514}  & 0.555          & \multicolumn{1}{c|}{0.363}     & 0.39     \\
\multicolumn{1}{c|}{}                               & Average         & \multicolumn{1}{c|}{0.607}          & 0.640          & \multicolumn{1}{c|}{0.626}  & \textbf{0.698} & \multicolumn{1}{c|}{0.599}  & 0.681          & \multicolumn{1}{c|}{0.389}     & 0.416    \\ \hline
\multicolumn{2}{c|}{Overall Average}                                                   & \multicolumn{1}{c|}{0.540}          & 0.595          & \multicolumn{1}{c|}{0.548}  & \textbf{0.617}          & \multicolumn{1}{c|}{0.536}  & \textbf{0.617} & \multicolumn{1}{c|}{0.353}     & 0.399    \\ \hline
\end{tabular}%
}
\caption{BLUCK benchmark comparison by subcategories and major categories across smaller models in 0-shot and 5-shot settings. The highest accuracy(s) for each category are boldy marked.}
\label{tab:per-cat-halfresults}
\end{table*}

\begin{figure*}[t]
  \centering
  \begin{subfigure}{0.8\linewidth}
    \includegraphics[width=\linewidth]{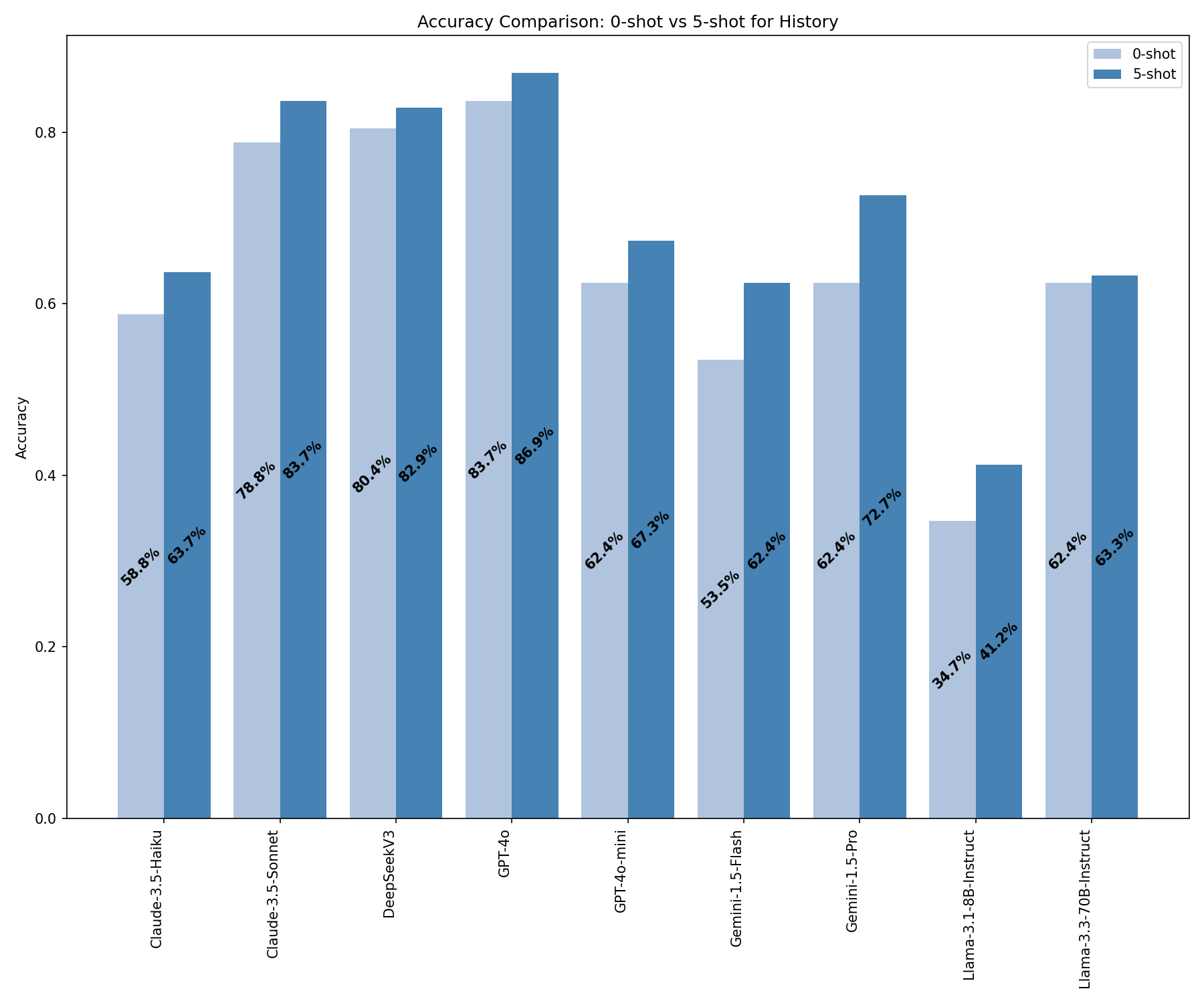}
    \caption{Accuracy for the history domain (0-shot and 5-shot).}
  \end{subfigure}
  \vspace{1em}
  
  \begin{subfigure}{0.8\linewidth}
    \includegraphics[width=\linewidth]{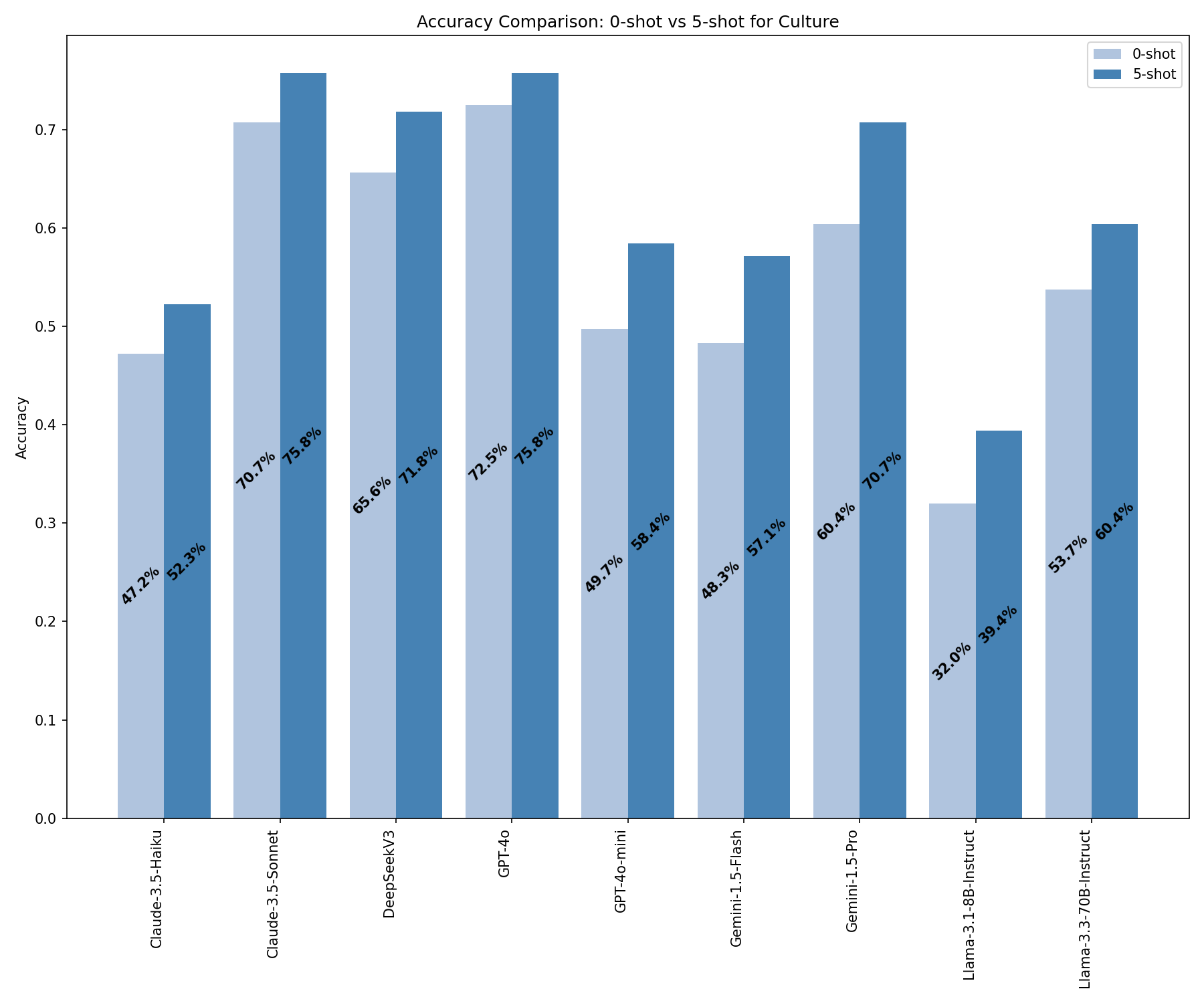}
    \caption{Accuracy for the culture domain (0-shot and 5-shot).}
  \end{subfigure}
  \caption{Comparison of accuracy across history and culture domains under 0-shot and 5-shot settings.}
  \vspace{0.5cm}
\end{figure*}
\begin{figure*}[t]
  \centering
  \begin{subfigure}{0.8\linewidth}
    \includegraphics[width=\linewidth]{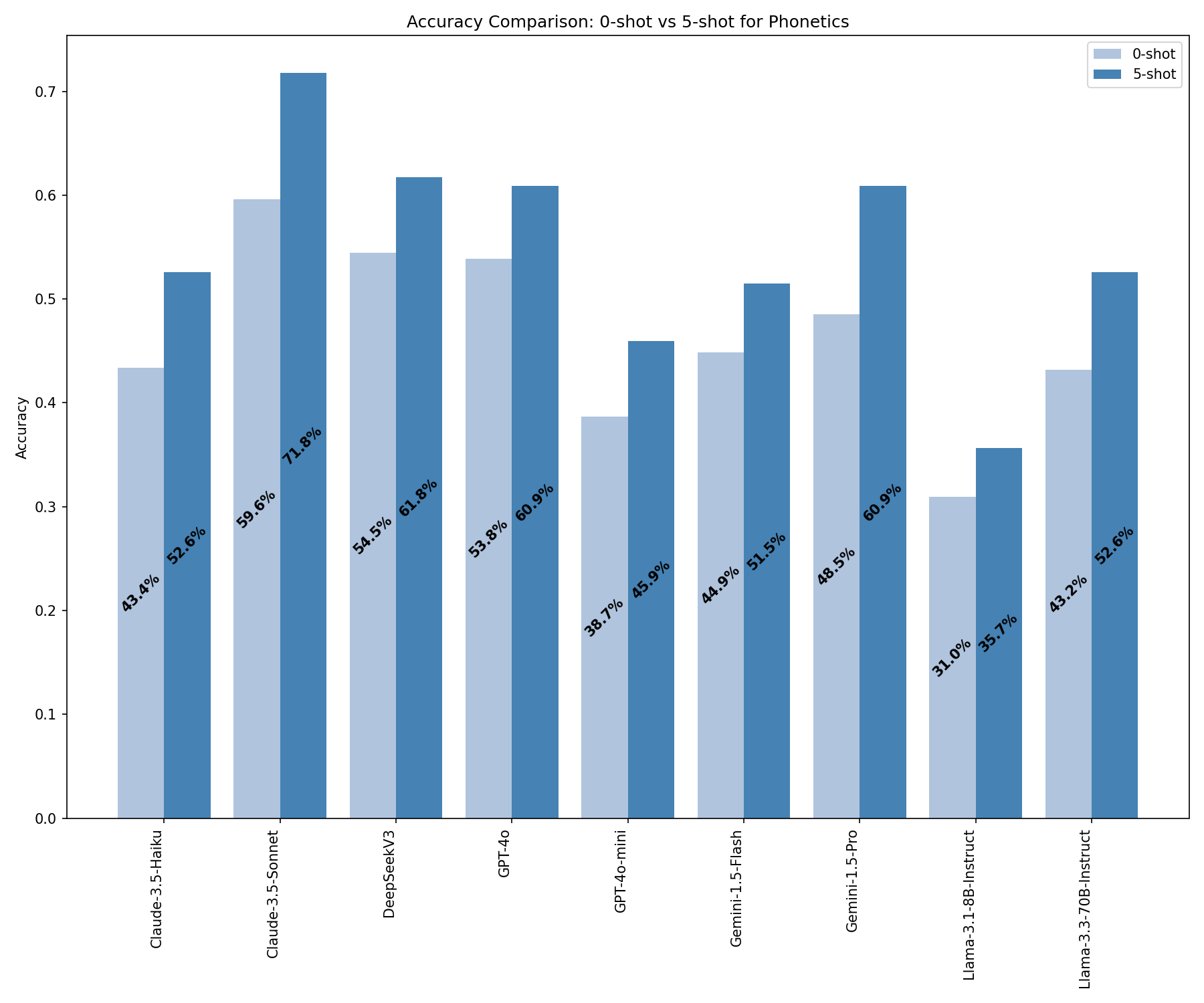}
    \caption{Accuracy for the phonetics domain (0-shot and 5-shot).}
  \end{subfigure}
  \vspace{1em}
  
  \begin{subfigure}{0.8\linewidth}
    \includegraphics[width=\linewidth]{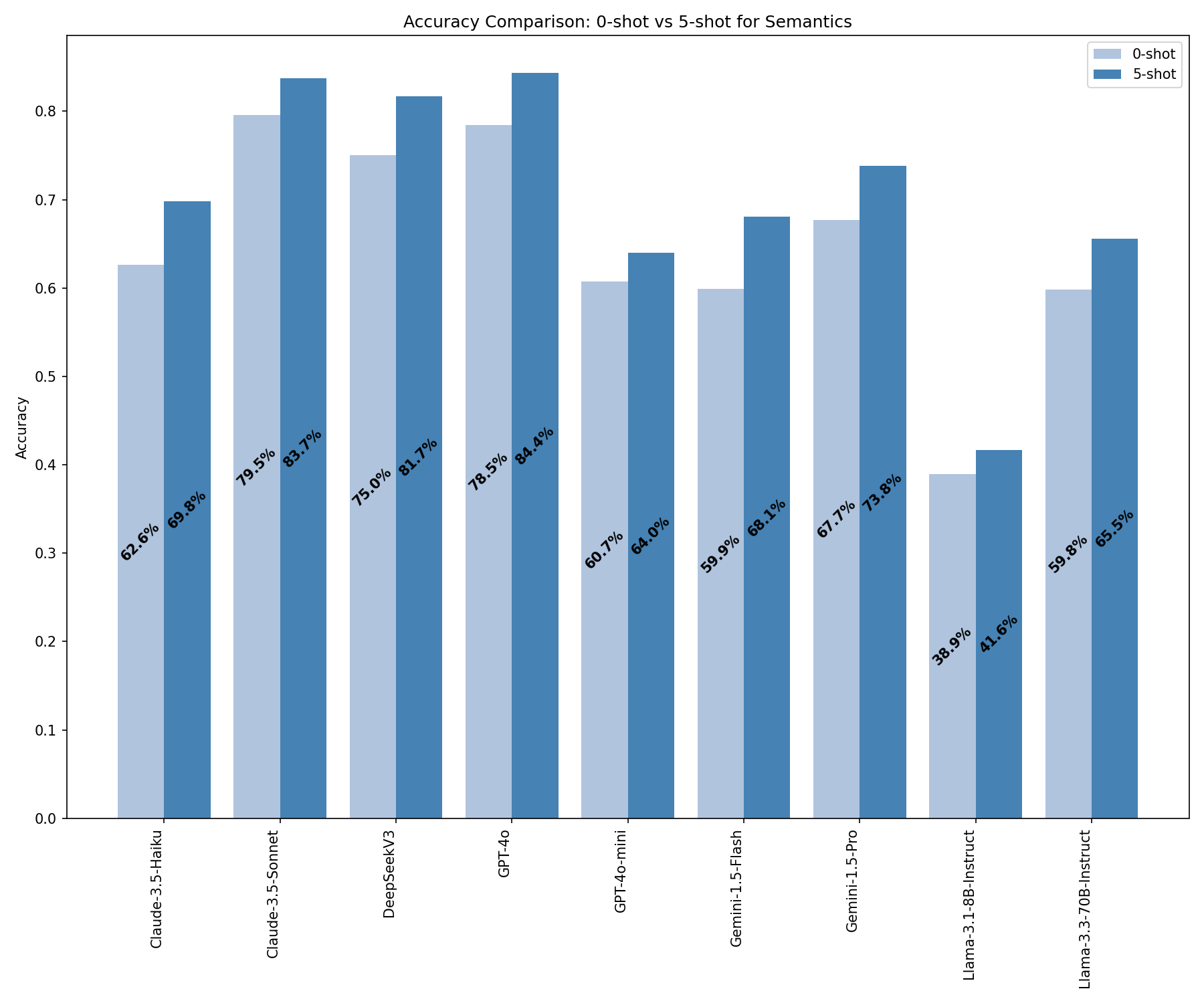}
    \caption{Accuracy for the semantics domain (0-shot and 5-shot).}
  \end{subfigure}
  
  \caption{Comparison of accuracy across phonetics and semantics domains under 0-shot and 5-shot settings.}
\end{figure*}

% \begin{figure*}[t]
%   \centering
  
%   % --- Figure 1 ---
%   % Set width to a fraction (e.g., 0.32\linewidth)
%   \begin{subfigure}{\linewidth}
%     \includegraphics[width=\linewidth]{Diagrams/Example_Semantics.png}
%     % \caption{Example MCQA from the Semantics domain.}
%   \end{subfigure}
%   % \hfill % Add horizontal space between 1 and 2
%   \caption{Example MCQA from the Semantics domain.}
% \end{figure*}

\begin{figure*}[t]
  \centering
  
  % --- Figure 2 ---
  \begin{subfigure}{\linewidth}
    \includegraphics[width=\linewidth]{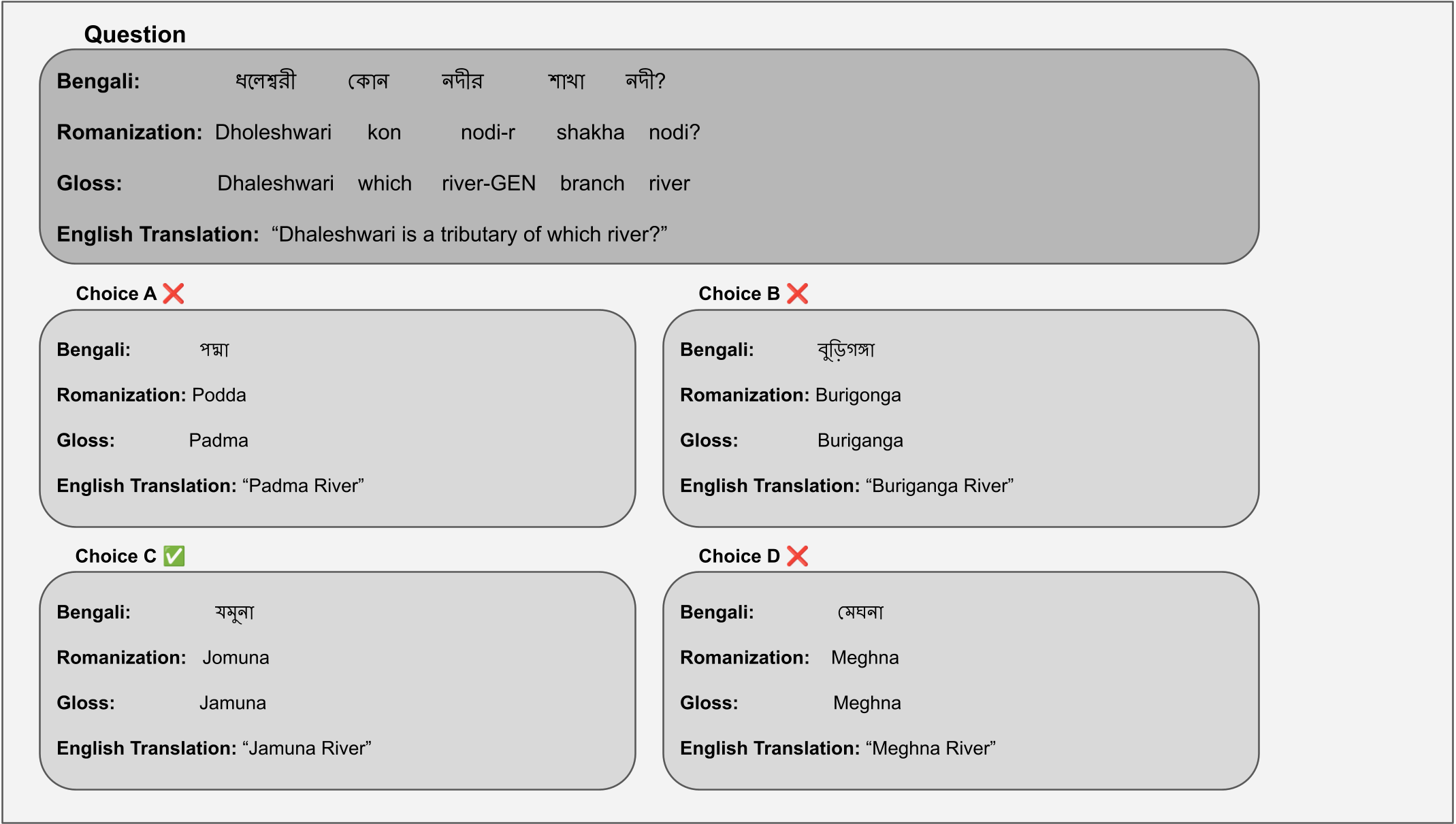}
  \end{subfigure}
  \caption{Example MCQA from the Culture domain.}
\end{figure*}

\begin{figure*}[t]
  \centering
  \begin{subfigure}{\linewidth}
    \includegraphics[width=\linewidth]{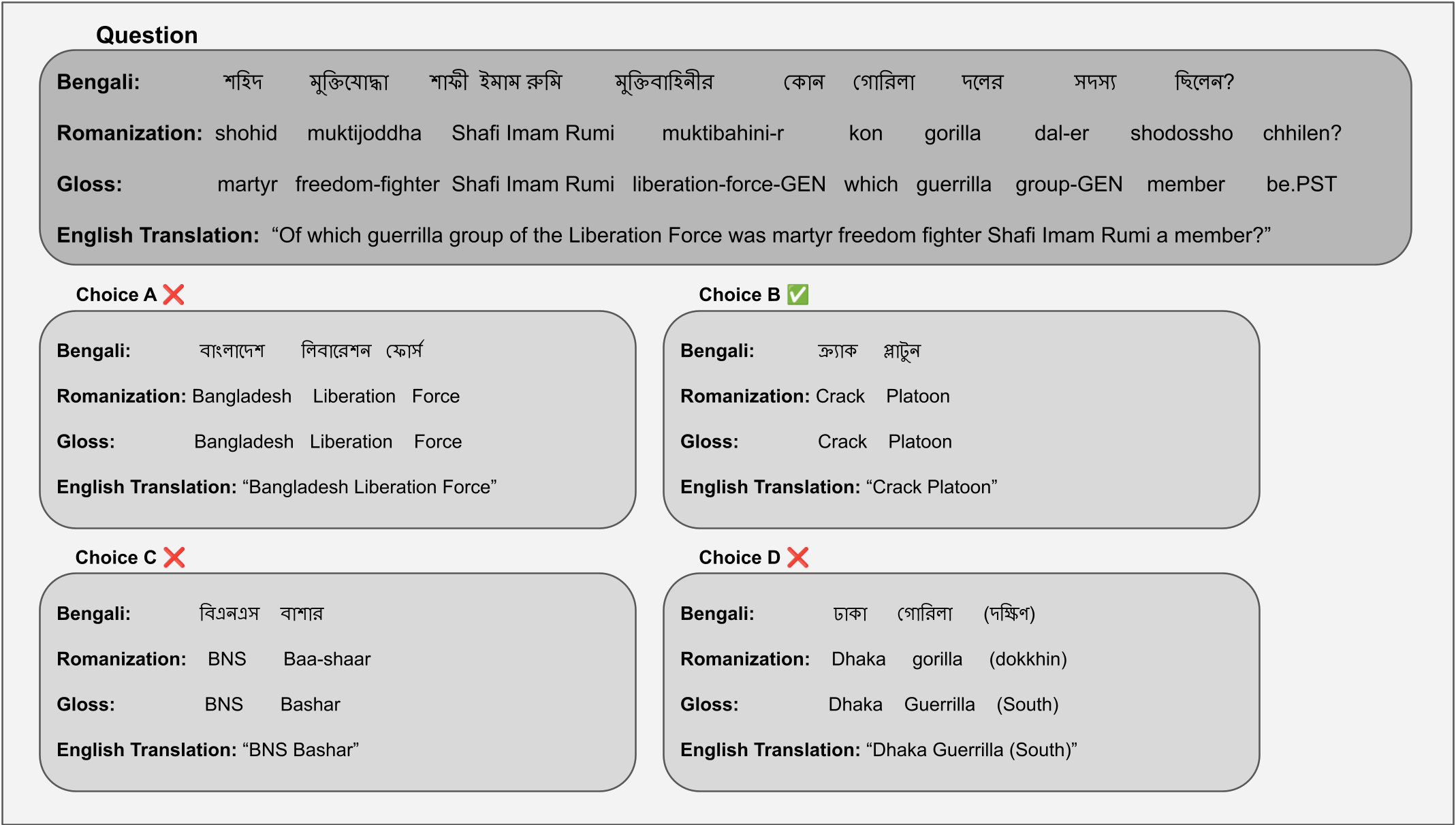}
  \end{subfigure}
  
  % REMOVED the stray \hfill from here
  \caption{Example MCQA from the History domain.}
\end{figure*}

\vspace{2mm}

\begin{figure*}[t]
  \centering
  \begin{subfigure}{\linewidth}
    \includegraphics[width=\linewidth]{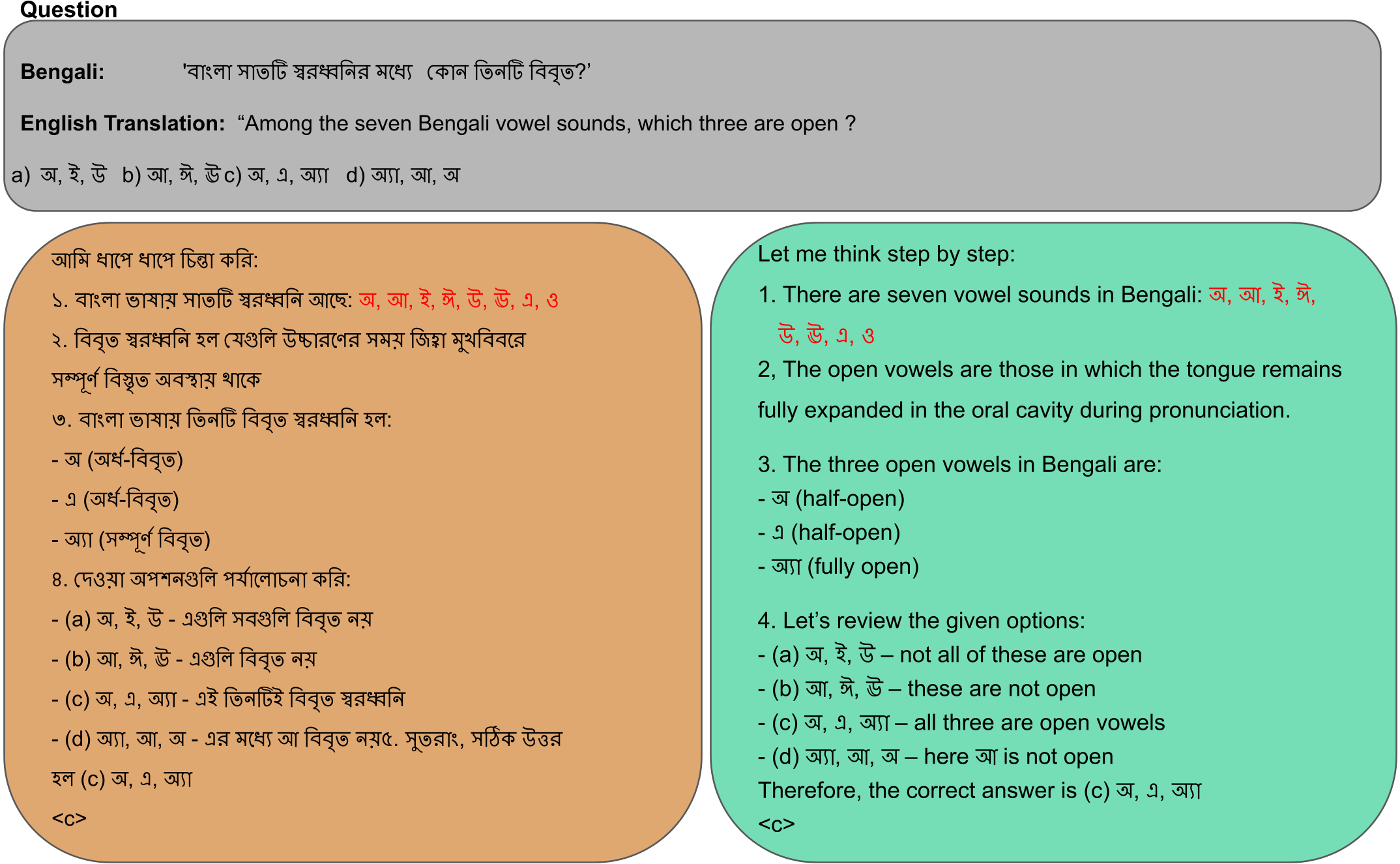}
    \caption{ In this case, the model fails (\textcolor{red}{red}) to acknowledge all seven correct Bengali vowel sounds in the beginning, which leads to erroneous reasoning.}
  \end{subfigure}
  
  \vspace{1em}
  \begin{subfigure}{\linewidth}
    \includegraphics[width=\linewidth]{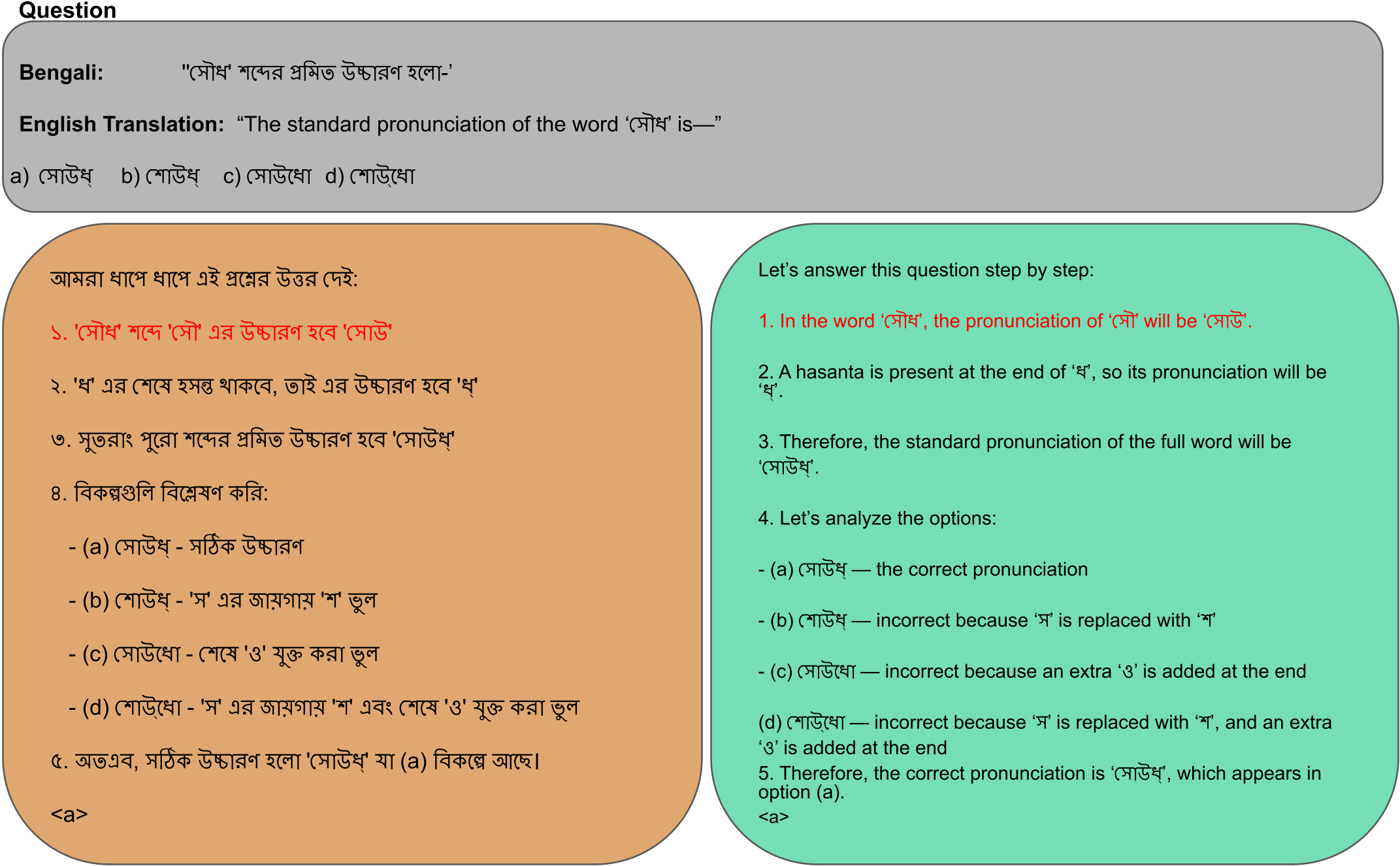}
    \caption{The model starts (\textcolor{red}{red}) with a hallucinated assumption of pronunciation of a certain Bengali letter and continues to explain on justifying the answer.}
  \end{subfigure}
  
  \caption{Failure Case in CoT Reasoning for Pronunciation Subcategory}
\end{figure*}

% \begin{figure*}[t]
%   \centering
%   \begin{subfigure}{\linewidth}
%     \includegraphics[width=\linewidth]{Diagrams/Example1_Pronunciation.png}
%   \end{subfigure}
%   \caption{Example MCQA from the Phonetics domain.}
% \end{figure*}

% \begin{figure*}[t]
%   \centering
%   \begin{subfigure}{\linewidth}
%     \includegraphics[width=\linewidth]{Diagrams/Example2_Pronunciation.png}
%   \end{subfigure}
%   \caption{Example MCQA from the Phonetics domain.}
% \end{figure*}

%   \caption{Data samples with Bengali Romanization and English gloss. The above applies for questions in the History, Culture, and Semantics categories.}
% \end{figure}

\end{document}